\pdfoutput=1

\documentclass[11pt]{article}

\usepackage[preprint]{acl}

\usepackage{graphicx}
\usepackage{subcaption}

\usepackage{times}
\usepackage{latexsym}

\usepackage[T1]{fontenc}

\usepackage[utf8]{inputenc}

\usepackage{microtype}

\usepackage{inconsolata}

\usepackage{graphicx}

\usepackage{float}

\usepackage{pgfplots}
\pgfplotsset{compat=newest}

\usepackage{booktabs}
\usepackage{array}

\title{Rethinking Thinking Tokens: \\Understanding Why They Underperform in Practice}

\author{
 \textbf{Sreeram Vennam\textsuperscript{1}},
 \textbf{David Valente\textsuperscript{2}},
 \textbf{David Herel\textsuperscript{3}},
 \textbf{Ponnurangam Kumaraguru\textsuperscript{1}},
\\
 \textsuperscript{1}IIIT Hyderabad \\
 \textsuperscript{2}Instituto Superior Técnico \\
 \textsuperscript{3}FEE, Czech Technical University in Prague,
\\
}

\begin{document}
\maketitle

\begin{abstract}

Thinking Tokens (TT) \cite{thinking-tokens} have been proposed as an unsupervised method to facilitate reasoning in language models.
However, despite their conceptual appeal, our findings show that TTs marginally improves performance and consistently underperforms compared to Chain-of-Thought (CoT) reasoning across multiple benchmarks.
We hypothesize that this underperformance stems from the reliance on a single embedding for TTs, which results in inconsistent learning signals and introduces noisy gradients.
This paper provides a comprehensive empirical analysis to validate this hypothesis and discusses the implications for future research on unsupervised reasoning in LLMs.

\end{abstract}

\section{Introduction}\label{sec:intro}

Large Language Models (LLMs) have demonstrated unprecedented performance across a wide array of natural language processing tasks, from translation to creative text generation. However, reasoning remains one of the key challenges in LLM research. Recent innovations, such as Chain-of-Thought (CoT) prompting \cite{cot-elicits-reasoning}, have shown considerable promise by breaking down complex tasks into sequential reasoning steps. This method has led to significant performance improvements on reasoning benchmarks like GSM8K and CommonsenseQA. The step-by-step nature of CoT enables explicit, interpretable reasoning but typically requires manual or supervised intervention through well-structured prompts.

Thinking Tokens (TTs) \cite{thinking-tokens}, in contrast, provide an unsupervised mechanism for reasoning. TTs work by introducing an intermediate ``thinking'' token, providing the model with more time to reason internally before generating output. This delay is theorized to allow deeper computation over hidden states, enhancing reasoning capacity by operating within the latent space of the model, which is potentially more expressive than the token space where CoT functions.

Despite their theoretical advantages, Thinking Tokens have underperformed in comparison to CoT across multiple reasoning benchmarks. This paper seeks to address two key questions: (1) How do Thinking Tokens compare against Chain-of-Thought prompting? (2) Why do Thinking Tokens underperform, and what could be a potential solution?

We hypothesize that TTs’ underperformance stems from their reliance on a single-token embedding, which may introduce noise and inconsistency during gradient updates. To explore this, we conduct controlled experiments on a range of reasoning tasks. Our contributions include: 1) An empirical comparison between TT and CoT reasoning across arithmetic, and symbolic tasks. 2) Identification of the root cause of TT’s underperformance. 3) Empirical validation of this hypothesis through gradient analysis.

\begin{figure*}[ht]
    \centering
    \includegraphics[width=0.8\linewidth]{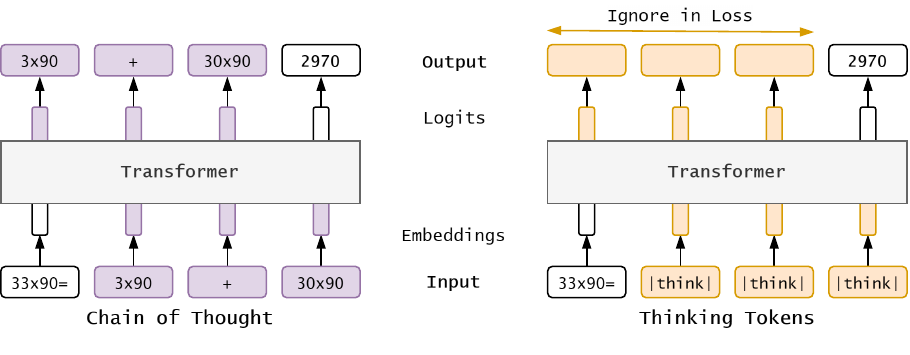}
    \caption{Chain of thought compared to thinking tokens. These approaches show striking similarity despite their differences.}
    \label{fig:tt-vs-cot}
\end{figure*}

\section{Related Work}\label{sec:related}

Reasoning in Large Language Models (LLMs) has been an active area of research in recent years. Chain-of-Thought (CoT) prompting \cite{wei2022chain} has emerged as a seminal technique, demonstrating that breaking down complex tasks into step-by-step reasoning improves performance across various benchmarks. CoT has proven particularly effective on tasks requiring multi-step reasoning, such as arithmetic problems, commonsense reasoning, and symbolic reasoning tasks \cite{zhang2023towards}.

To mitigate the need for manually structured prompts, Thinking Tokens (TTs) \cite{thinking-tokens} were introduced as an unsupervised approach to reasoning. TTs attempt to extend the reasoning capabilities of LLMs by inserting a ``thinking'' token, allowing the model more time to compute over latent states before output generation. While this method promises deeper internal computation, empirical results show that TTs fall short of the performance achieved by CoT, likely due to training instabilities introduced by a single-token embedding. In parallel, other methods like Pause Tokens \cite{pause-tokens} have been explored to simulate cognitive pauses, placing tokens at random intervals to mimic the effect of reasoning time. However, none of these alternatives have matched the structured reasoning power that CoT delivers. Moreover, studies have delved into understanding the nature of reasoning in LLMs by analyzing their circuit complexity, and internal representations, offering insights into how models handle tasks requiring logical and sequential thinking \cite{zhang2023towards, length-complexity}. These analyses are crucial for understanding why approaches like CoT excel, while unsupervised methods like Thinking Tokens struggle to achieve similar success.

\section{Hypothesis and Theoretical Analysis}
\label{sec:hypothesis}

We hypothesize that the core issue with Thinking Tokens lies in the embedding mechanism. When a single embedding is used for TTs, during backpropagation the model receives inconsistent learning signals, leading to noisy gradient updates. This noise disrupts learning, particularly in tasks that require structured intermediate steps, such as arithmetic reasoning or multi-hop commonsense tasks.

To formalize this, let $\mathbf{h}_t$ represent the hidden state at time step $t$, and let $\mathbf{e}_{TT}$ be the embedding associated with the Thinking Token. The gradient $\nabla L(\mathbf{e}_{TT})$ with respect to the loss function $L$ becomes inconsistent across training examples, as $\mathbf{e}_{TT}$ serves multiple, contextually distinct roles across the token sequence. This contrasts with CoT, where each reasoning step has an explicit, interpretable role in the output.




\begin{table*}[ht]
\centering
\begin{tabular}{
    l
    |
    >{\centering\arraybackslash}p{0.8cm} 
    >{\centering\arraybackslash}p{0.8cm} 
    >{\centering\arraybackslash}p{0.8cm} 
    | 
    >{\centering\arraybackslash}p{1.8cm} 
    >{\centering\arraybackslash}p{1.8cm}
}
\toprule
& \multicolumn{3}{c}{\textbf{Digit Multiplication}} & \multicolumn{2}{c}{\textbf{Natural Language Datasets}} \\
\cmidrule(lr){2-4} \cmidrule(lr){5-6}
& 2d & 3d & 4d & GSM8k & OpenBookQA \\
\midrule
Baseline    & $0.0$ & $0.0$ & $0.0$ & $6.30$ & $37.2$ \\
TT          & $7.32$ & $0.01$ & $0.0$ & $4.51$ & $37.2$ \\
\midrule
CoT         & $91.9$ & $66.3$ & $>0$ & $18.7$ & $42.0$ \\
TT + CoT    & $92.3$ & $67.8$ & $>0$ & $17.5$ & $39.6$ \\
\bottomrule
\end{tabular}
\caption{\centering Performance of our four configurations across various tasks. We report accuracy over the integer value in Digit Multiplication. We report exact match accuracy on GSM8k and OpenBookQA. $>0$ indicates the experiment produced non-zero results but was killed early due to resource constraints.}
\label{tab:placeholder-values}
\end{table*}

\subsection{Embedding Space, Gradient Dynamics, and Chain Representation}

The core principles behind both Thinking Tokens (TTs) and Chain-of-Thought (CoT) lie in how they facilitate intermediate reasoning steps within a model. While CoT provides explicit intermediate representations, TTs introduce a single or multiple shared token embeddings to emulate intermediate steps. In CoT, reasoning steps are encoded as distinct tokens \( e_{CoT}^1, e_{CoT}^2, \dots, e_{CoT}^m \), where \( m \) denotes the number of reasoning steps. The transformer model generates a sequence of hidden states:
\begin{equation}
    h_t = f(e_{CoT}^t, h_{t-1}),
\end{equation}
where \( h_{t-1} \) is the hidden state from the previous step, and \( f \) is a non-linear transformation (e.g., a multi-layer perceptron). The explicit decomposition of reasoning into distinct tokens yields structured and stable gradient updates:
\begin{equation}
    \nabla L(e_{CoT}^t) = \frac{\partial L}{\partial h_t},
\end{equation}
where \( L \) is the loss function. Each \( e_{CoT}^t \) represents a unique reasoning subtask, allowing the model to isolate learning signals for each step and reduce noise during training.

Thinking Tokens (TTs) adopt a different mechanism by using one or more shared tokens \( e_{TT}^1, e_{TT}^2, \ldots, e_{TT}^k \) to facilitate internal reasoning:
\begin{equation}
    h_t = f(e_{TT}^i, h_{t-1}),
\end{equation}
where \( i \) indexes the set of shared tokens. In this case, the gradient updates are noisier because the same token embeddings \( e_{TT}^i \) are reused across multiple reasoning steps:
\begin{equation}
    \Delta e_{TT} = \sum_{i=1}^{k} \nabla L(e_{TT}^i).
\end{equation}
This reuse of embeddings across different contexts introduces ambiguity into the learning signals, making it harder for the model to cleanly separate the contribution of each reasoning step, leading to a noisier gradient signal compared to CoT.

\begin{figure}[h]
    \centering
    \includegraphics[width=0.9\linewidth, height=5cm]{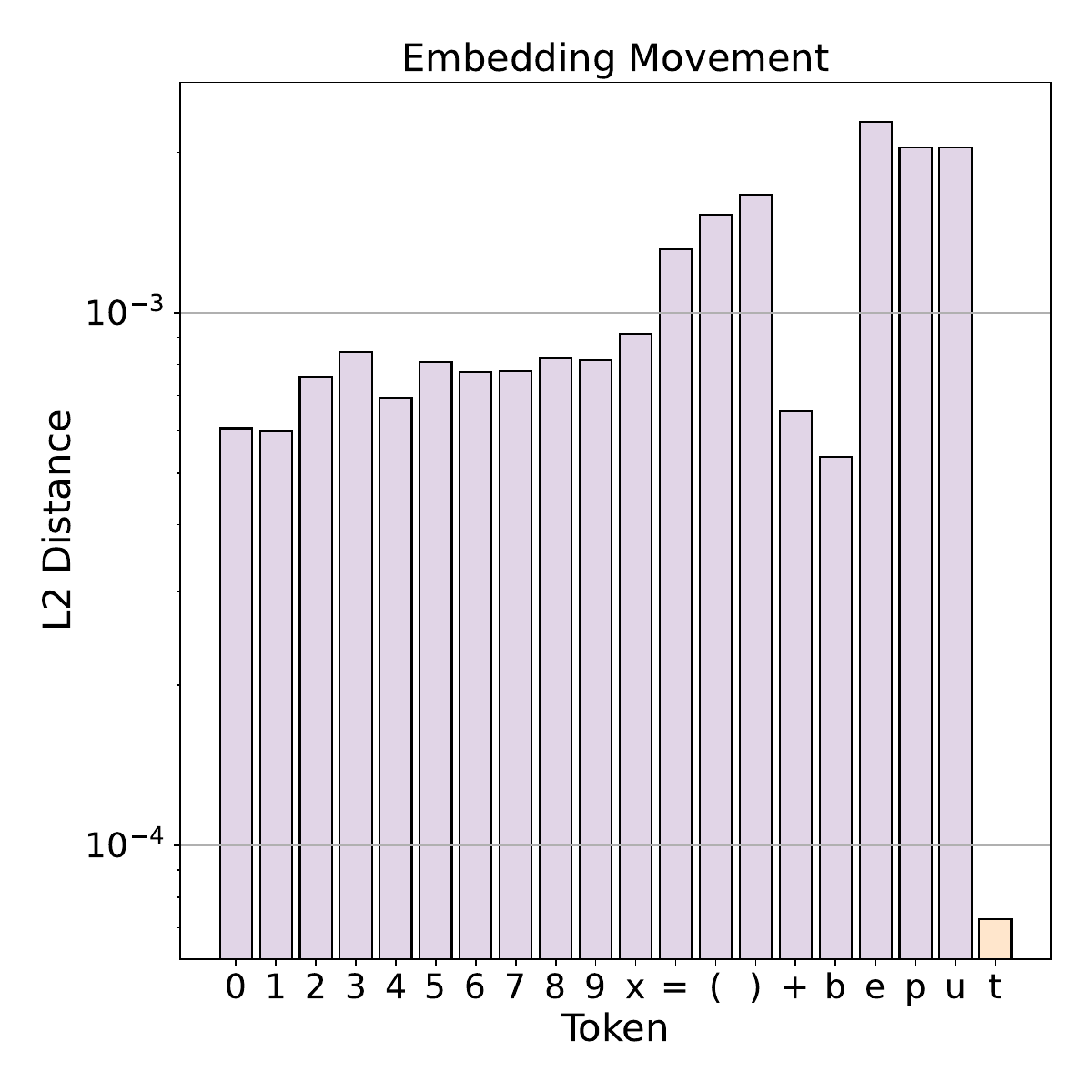}
    \caption{One TT embedding hardly moves from the initialized value.}
    \label{fig:one-tt-embedding}
\end{figure}

\begin{figure}
    \centering
    \includegraphics[width=0.9\linewidth, height=5cm]{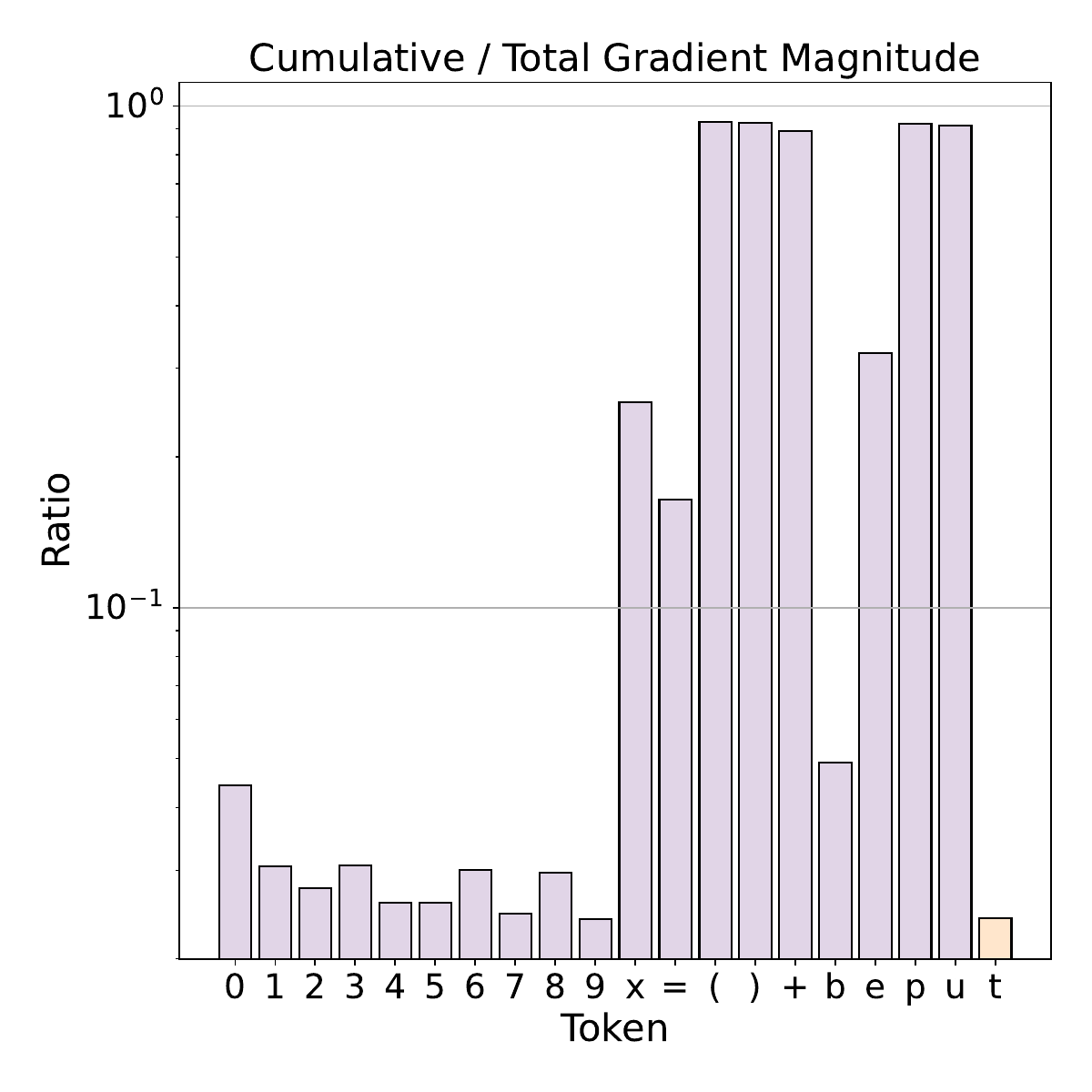}
    \caption{One TT embedding receives insufficient cumulative gradient.}
    \label{fig:one-tt-cumulative}
\end{figure}

\section{Experiments}\label{sec:experiments}

We first compare TT against CoT in section~\ref{tt-cot} on both synthetic data and popular natural language benchmarks. We perform our analysis on tasks in which intermediate computation can help neural networks learn, such as GSM8k. We then verify our hypothesis empirically in section~\ref{sec:grad-analysis} by monitoring both the embedding weights and the gradient of the vocabulary layer of the model.

\subsection{TT vs CoT}\label{tt-cot}

We compare performance across the below 4 configurations. Our results can be found in Table~\ref{tab:placeholder-values}.

\paragraph{Baseline} The base configuration without intermediate steps, expecting the multiplication result immediately after the expression.

\paragraph{CoT} Pretraining on CoT using the intermediate calculations. This involves supervising the model with the intermediate calculations during training.

\paragraph{TT} Pretraining using TTs hoping that the model learns to use these tokens to reason through the problem.

\paragraph{TT + CoT} Combining CoT with TT during pretraining.

\subsubsection{Digit Multiplication}

We adopt a synthetic dataset for integer multiplication from \cite{length-complexity} which includes intermediate calculations that takes the role of CoT. We conduct our experiments using four main configurations across 2-digit, 3-digit, and 4-digit multiplication. The think token is depicted as \textit{"t"}.


    
    


\paragraph{Experimental Setup}{
We train a GPT-2 based transformer for 100 epochs, utilizing proportionally more data samples as the digit count increased. Implementation details can be found in the Appendix.
}


\subsubsection{Natural Language Tasks}

CoT can aid NLP tasks as well. In this section, we compare TT against CoT for standard NLP tasks such as mathematical reasoning (GSM 8k) and question answering (OpenBookQA).

\paragraph{Experimental Setup}{
We finetune Llama 3.2 (1B) \cite{llama-3-family} on two datasets, GSM8k, and OpenBookQA. CoT data for these datasets are available through ThoughtSource~\cite{https://doi.org/10.48550/arxiv.2301.11596}. Implementation details can be found in the Appendix.
}


\subsection{Gradient Analysis}\label{sec:grad-analysis}

To validate our hypothesis (Section~\ref{sec:hypothesis}) empirically, we record the embeddings and gradients of the embedding layer regularly during training and analyze them. Furthermore, we introduce two thinking tokens with distinct embeddings, \textit{``t''} and \textit{``ts''} and monitor their embeddings as well, if our hypothesis is true, we should see clearer gradients using two distinct tokens.

\paragraph{Noisy Gradients} We calculate how far the embedding travels. If the embedding hasn't moved much, during training, that could be a strong indicator of noisy gradients. We also calculate the cumulative gradient. Noisy gradients result in a low cumulative gradient since the mean of noise is zero.

\paragraph{Inconsistent Learning Signals} We calculate the gradient direction variance -- high directional variance where the token embedding erratically changes to accommodate every context can be a sign of inconsistent learning signals.

\begin{figure}
    \centering
    \includegraphics[width=0.9\linewidth, height=5cm]{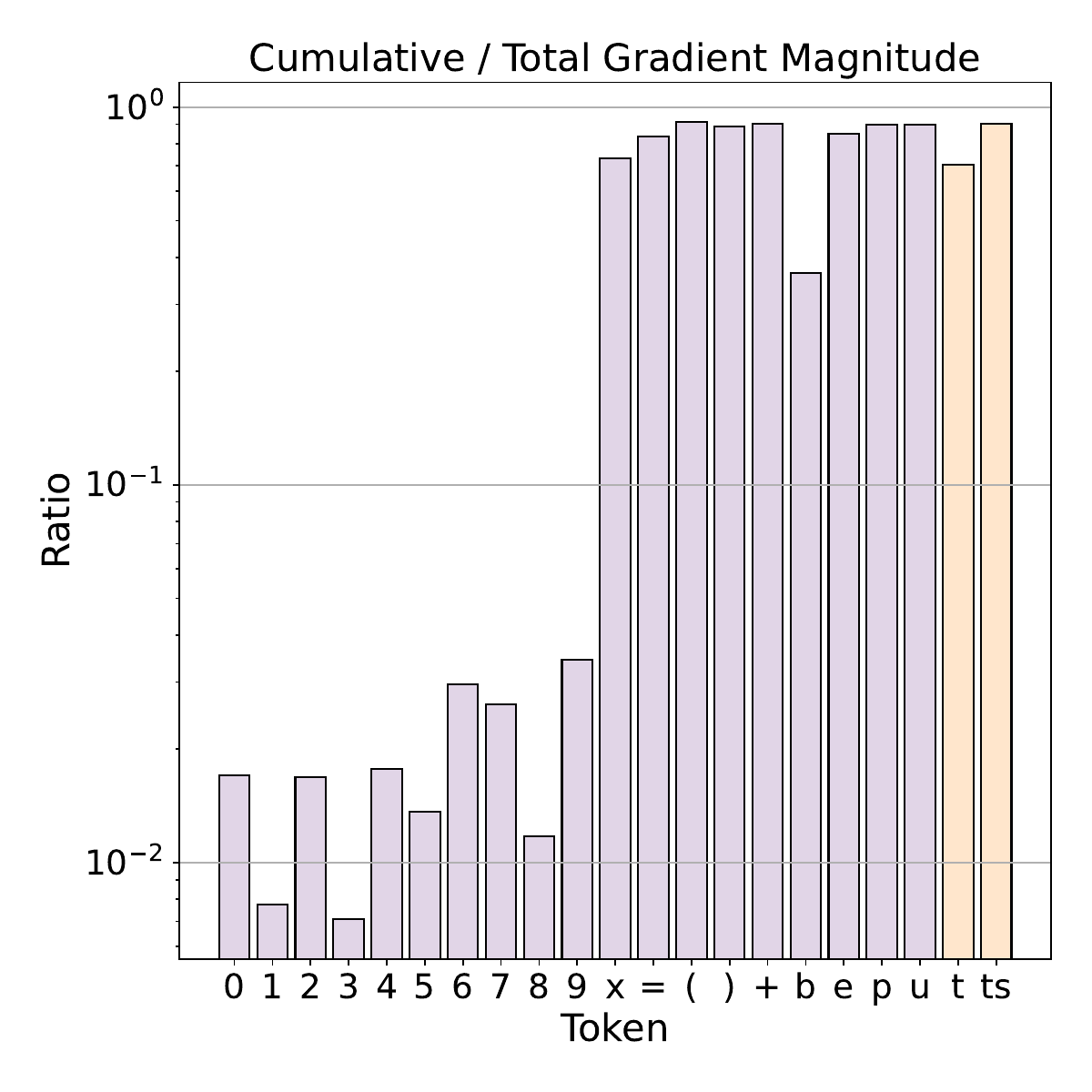}
    \caption{Two TT embeddings receive clear large cumulative gradients.}
    \label{fig:two-tt-cumulative-gradients}
\end{figure}

\begin{figure}
    \centering
    \includegraphics[width=0.9\linewidth, height=5cm]{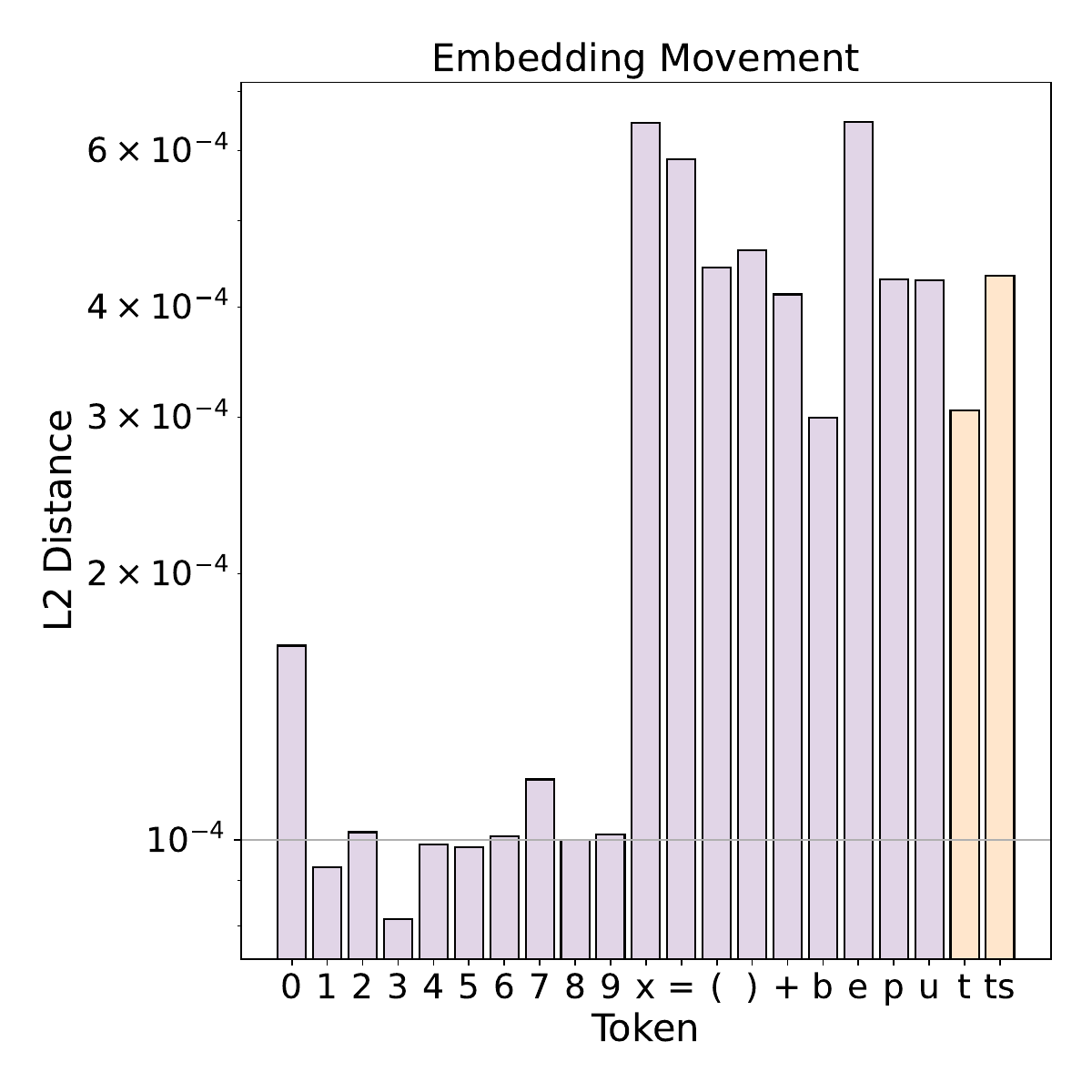}
    \caption{Two TT embeddings show clear deviation from initialization.}
    \label{fig:two-tt-gradient-variance}
\end{figure}


\section{Analysis}\label{sec:analysis}

Table~\ref{tab:placeholder-values} shows us that TT performs worse that CoT across multiple benchmarks. Sometimes it can even hurt model performance, a finding echoed by \cite{pause-tokens}. In summary, TT offers only marginal gains while being overshadowed by CoT.

\subsection{Gradient Analysis}

\paragraph{One Embedding} We see in Figure~\ref{fig:one-tt-embedding} that the embedding hardly travels from its initialization showing the tokens limited expressivity in practice. Figure~\ref{fig:one-tt-cumulative} describes how the cumulative gradient is the smallest of the tokens explain why the embedding is stagnant.

\paragraph{Two Embeddings} The addition of two unique embeddings results in significantly clearer gradients. Figure~\ref{fig:two-tt-cumulative-gradients} shows that these tokens now receive solid gradients. Figure~\ref{fig:two-tt-gradient-variance} reveals that the embeddings are no longer stagnant and move from initialization, verifying our hypothesis.

\section{Discussion}\label{sec:discussion}
Our results suggest that while Thinking Tokens offer a novel unsupervised mechanism for reasoning, they suffer from noisy gradients due to their single embedding mechanism. When paired with CoT, they perform as if they weren't present. Although TT looks appealing due to unsupervised reasoning within the latent space, future approaches should aim to better make use of this space with richer vectors rather than a single embedding.






\clearpage
\bibliography{references}

\begin{thebibliography}{8}
\providecommand{\natexlab}[1]{#1}

\bibitem[{Dubey et~al.(2024)Dubey, Jauhri, Pandey, Kadian, Al-Dahle, Letman, Mathur, Schelten, Yang, Fan et~al.}]{llama-3-family}
Abhimanyu Dubey, Abhinav Jauhri, Abhinav Pandey, Abhishek Kadian, Ahmad Al-Dahle, Aiesha Letman, Akhil Mathur, Alan Schelten, Amy Yang, Angela Fan, et~al. 2024.
\newblock The llama 3 herd of models.
\newblock \emph{arXiv preprint arXiv:2407.21783}.

\bibitem[{Goyal et~al.(2024)Goyal, Ji, Rawat, Menon, Kumar, and Nagarajan}]{pause-tokens}
Sachin Goyal, Ziwei Ji, Ankit~Singh Rawat, Aditya~Krishna Menon, Sanjiv Kumar, and Vaishnavh Nagarajan. 2024.
\newblock \href {https://openreview.net/forum?id=oBpHVLzkfm} {Think before you speak: Training language models with pause tokens}.
\newblock In \emph{International Conference on Learning Representations (ICLR)}.

\bibitem[{Herel and Mikolov(2023)}]{thinking-tokens}
David Herel and Tomas Mikolov. 2023.
\newblock \href {https://arxiv.org/abs/2405.08644} {Thinking tokens for language modeling}.
\newblock In \emph{Proceedings of the 8th Conference on Artificial Intelligence and Theorem Proving (AITP 2023)}.

\bibitem[{Malach et~al.(2024)}]{length-complexity}
Eran Malach et~al. 2024.
\newblock \href {https://icml.cc/virtual/2024/poster/33369} {Auto-regressive next-token predictors are universal learners}.
\newblock In \emph{Proceedings of the 41st International Conference on Machine Learning (ICML)}.

\bibitem[{Ott et~al.(2023)Ott, Hebenstreit, Liévin, Hother, Moradi, Mayrhauser, Praas, Winther, and Samwald}]{https://doi.org/10.48550/arxiv.2301.11596}
Simon Ott, Konstantin Hebenstreit, Valentin Liévin, Christoffer~Egeberg Hother, Milad Moradi, Maximilian Mayrhauser, Robert Praas, Ole Winther, and Matthias Samwald. 2023.
\newblock \href {https://doi.org/10.48550/ARXIV.2301.11596} {Thoughtsource: A central hub for large language model reasoning data}.
\newblock \emph{arXiv preprint}.

\bibitem[{Wei et~al.(2022{\natexlab{a}})Wei, Wang, Schuurmans, Bosma, ichter, Xia, Chi, Le, and Zhou}]{cot-elicits-reasoning}
Jason Wei, Xuezhi Wang, Dale Schuurmans, Maarten Bosma, brian ichter, Fei Xia, Ed~Chi, Quoc~V Le, and Denny Zhou. 2022{\natexlab{a}}.
\newblock \href {https://proceedings.neurips.cc/paper_files/paper/2022/file/9d5609613524ecf4f15af0f7b31abca4-Paper-Conference.pdf} {Chain-of-thought prompting elicits reasoning in large language models}.
\newblock In \emph{Advances in Neural Information Processing Systems}, volume~35, pages 24824--24837. Curran Associates, Inc.

\bibitem[{Wei et~al.(2022{\natexlab{b}})}]{wei2022chain}
Jason Wei et~al. 2022{\natexlab{b}}.
\newblock Chain of thought prompting elicits reasoning in large language models.
\newblock \emph{arXiv preprint arXiv:2201.11903}.

\bibitem[{Zhang et~al.(2023)Zhang, Sun et~al.}]{zhang2023towards}
Xue Zhang, Yuchen Sun, et~al. 2023.
\newblock Towards revealing the mystery behind chain of thought: A theoretical perspective.
\newblock \emph{arXiv preprint arXiv:2305.15408}.

\end{thebibliography}

\clearpage
\appendix
\section{Appendix}\label{sec:appendix}

\subsection{Experimental Setup}

\subsubsection{Digit Multiplication Experiments}
For our digit multiplication experiments, we utilized a synthetic dataset for integer multiplication as described in Malach et al. (2024). The following configurations were employed during the training process:

\begin{itemize}
    \item \textbf{Model:} We implemented a GPT-2 based transformer model.
    \item \textbf{Epochs:} The model was trained for a total of 50 epochs.
    \item \textbf{Batch Size:} A batch size of 128 was used throughout the training.
    \item \textbf{Learning Rate:} The learning rate was determined using a learning rate finder as implemented in PyTorch Lightning (cite as lr-finder).
    \item \textbf{Hardware:} The experiments were conducted on 2 L40s (GPUs).
\end{itemize}
The dataset included multiple samples across different digit counts, specifically targeting 2-digit, 3-digit, and 4-digit multiplication tasks.

\subsubsection{Natural Language Processing Tasks}
For the NLP tasks, we focused on fine-tuning the Llama 3.2 (1B) model on established benchmarks:

\begin{itemize}
    \item \textbf{Datasets:} The models were evaluated on GSM8K and OpenBookQA datasets, which are known for their reasoning complexity.
    \item \textbf{Epochs:} The training was performed for 5 epochs.
    \item \textbf{Batch Size:} An effective batch size of 512 was employed to ensure effective training dynamics.
    \item \textbf{Hardware:} The experiments were carried out using 2 L40s (GPUs).
\end{itemize}
The experiments aimed to assess the performance of Thinking Tokens (TTs) compared to Chain-of-Thought (CoT) prompting in various reasoning tasks, leveraging the structured reasoning benefits offered by CoT.

\end{document}